\documentclass[runningheads]{llncs}
\usepackage{graphicx}
\usepackage{color}
\usepackage{graphicx}
\usepackage{amsmath}
\usepackage{amssymb}
\usepackage{booktabs}
\newcommand*{\rom}[1]{\expandafter\@slowromancap\romannumeral #1@}

\usepackage[breaklinks=true,bookmarks=false]{hyperref}
\begin{document}
\title{Automatic Operating Room Surgical Activity Recognition for Robot-Assisted Surgery}
%
%
\author{Aidean Sharghi, Helene Haugerud, Daniel Oh, Omid Mohareri}
%
\authorrunning{}
%
\institute{Intuitive Surgical Inc., Sunnyvale, CA}
\maketitle              
\vspace{-15pt}
\begin{abstract}
Automatic recognition of surgical activities in the operating room (OR) is a key technology for creating next generation intelligent surgical devices and workflow monitoring/support systems. Such systems can potentially enhance efficiency in the OR, resulting in lower costs and improved care delivery to the patients. In this paper, we investigate automatic surgical activity recognition in robot-assisted operations. We collect the first large-scale dataset including 400 full-length multi-perspective videos from a variety of robotic surgery cases captured using Time-of-Flight cameras. We densely annotate the videos with 10 most recognized and clinically relevant classes of activities. Furthermore, we investigate state-of-the-art computer vision action recognition techniques and adapt them for the OR environment and the dataset. First, we fine-tune the Inflated 3D ConvNet (I3D) for clip-level activity recognition on our dataset and use it to extract features from the videos. These features are then fed to a stack of 3 Temporal Gaussian Mixture layers which extracts context from neighboring clips, and eventually go through a Long Short Term Memory network to learn the order of activities in full-length videos. We extensively assess the model and reach a peak performance of ${\raise.17ex\hbox{$\scriptstyle\sim$}} 88\%$ mean Average Precision.
\end{abstract}
\vspace{-25pt}
\section{Introduction}
\vspace{-10pt}
Robot-assisted surgery (RAS) has been shown to improve perioperative outcomes such as reduced blood loss, faster recovery times, and shorter hospital stays for certain procedures. However, cost, staff training requirements and increased OR workflow complexities can all be considered barriers to adoption~\cite{sel}. Although human error in care delivery is inevitable in such complex environments, there is opportunity for data-driven workflow analysis systems to identify “error-prone” situations, anticipate failures and improve team efficiency in the OR~\cite{c4c}. To this end, there have been a few observational studies to analyze non-operative activities, robot setup workflows and team efficiency aiming to enhance total system performance in robotic surgery~\cite{sel,eiw,fao}. However, such approaches are time and cost intensive and are not scalable.

\begin{figure*}[t]
	\centering
	\includegraphics[width=\linewidth]{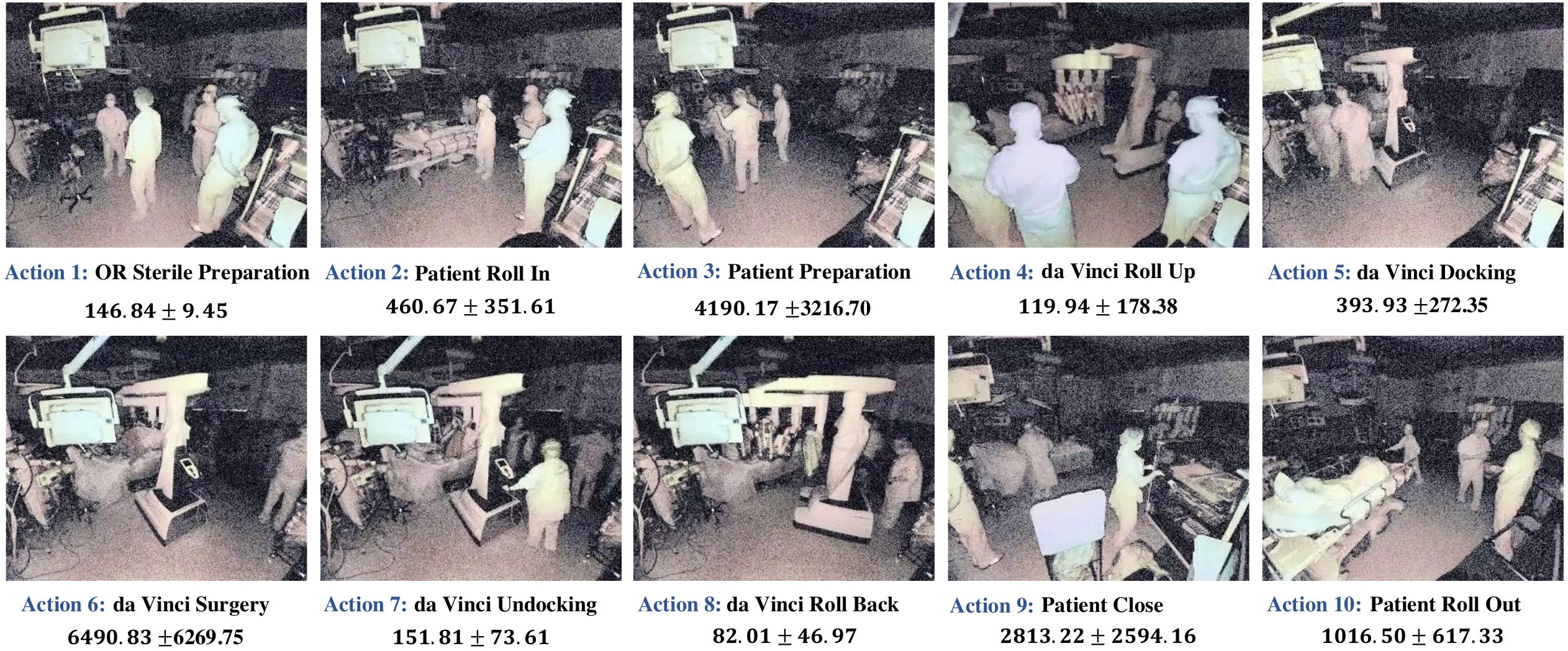}
	\caption{\scriptsize We annotate each video in the dataset with its constituent actions (activities). Generally, 10 activities take place in every video. We report mean and standard deviation of number of frames per activity under the samples in this figure.}
	\label{fig:bar} 
	\vspace{-20pt}
\end{figure*}

\begin{figure*}[t]
	\centering
	\includegraphics[width=0.8\linewidth]{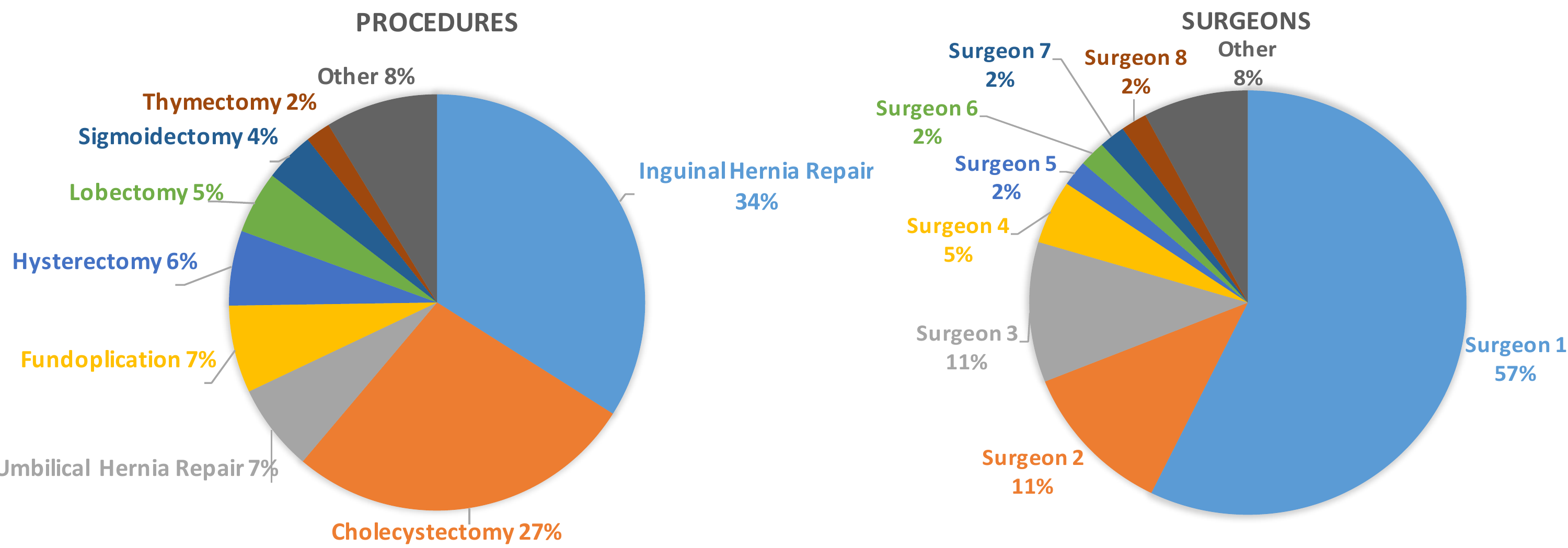}
	\caption{\scriptsize A breakdown of the videos in our dataset based on type of procedure and surgeons.}
	\label{fig:surg} 
	\vspace{-15pt}
\end{figure*}

In this paper, we aim to develop a framework to automatically recognize the activity that is taking place at any moment in a full-length video captured in an OR. Almost all state-of-the-art approaches to activity recognition are based on data-driven approaches (in machine learning and computer vision), i.e., models are trained on large-scale datasets. Hence, in order to take advantage of them, we collect a \textit{large-scale robot-assisted operation} dataset, including 400 full-length videos captured from 103 surgical cases performed by the da Vinci Xi surgical system. This is achieved by placing two vision carts, each with two Time-of-Flight (ToF) cameras inside operating rooms to create a multi-view video dataset. The reason behind using ToF cameras instead of high resolution RGB cameras is twofold: 1) to capture 3D information about the scene, 2) to preserve the privacy of patients and staff involved. Our dataset includes cases from several types of operations (see Figure~\ref{fig:surg}). After collecting the data, a trained user annotated each video with 10 clinically relevant activities, illustrated by Figure~\ref{fig:bar}. 

Next, we develop an effective method to perform surgical activity recognition, i.e., automatically identifying activities that occur at any moment in the video. This problem is different from existing action recognition studied in the computer vision community. Firstly, our activities of interest have high inter/intra-class length variations. Secondly, surgical activities follow an order; i.e., first a sterile table is set up with sterile instruments (sterile preparation), then the patient is rolled in and prepared, etc. If the order is not accounted for, existing state-of-the-art methods fail to reach reliable performance due to inherent resemblance between certain activities such as robot docking and undocking. Thirdly, the videos in our dataset are full-length surgery cases and some last over 2 hours. Effective learning on such long videos is difficult and precautions must be taken.

Given a video, we first partition it uniformly into short clips (16 consecutive frames), and then feed them to the model in sequential order. Once the model has seen all the clips, it predicts the class of activities for every clip in the video. The right panel of Figure~\ref{fig:app} shows our proposed framework. Our model consists of two sub-networks. The first sub-network, shown on the left in Figure~\ref{fig:app}, is the Inflated 3D ConvNet (I3D)~\cite{i3d} that is commonly used to extract discriminative features from the videos. 
We first fine-tune I3D (pretrained on ImageNet) on our dataset, and then use it to extract spatio-temporal features for the videos. Next, we employ 3 Temporal Gaussian Mixture (TGM)~\cite{tgm} layers, each with several per-class temporal filters. These layers help with extraction of features from neighboring clips, enabling the model to eventually classify each clip with contextual knowledge. Finally, we process the clips in temporal order using a Long Short Term Memory (LSTM)~\cite{lstm} to learn the order of activities performed. We comprehensively assess the generalization capability of our approach by splitting our videos into train and test according to different criteria, and are able to reach peak performance as high as $\sim88\%$ mean Average Precision (mAP).

When dealing with robot-assisted surgeries, event and kinematics data generated by the surgical robot can be used to identify certain activities. However, this cannot replace the need for additional sensing as other activities do not involve interactions with the robot (e.g., patient in/out etc.). A smart OR integration depends on reliable detection of such activities. While in this work we study robot-assisted surgeries, detecting activities in the OR is still important in open and laparoscopic cases where such system data is not available. By not relying on such meta-data, we develop a more general solution to the problem.

\vspace{-15pt}
\section{Related Work}
\vspace{-10pt}
Automatic activity recognition has gained a lot of attention recently.
Depth sensors are used in hospitals to monitor hand hygiene compliance using CNNs~\cite{hhr} and detect patient mobility~\cite{mpm,dld}. In~\cite{ppa}, a super-resolution model is used to enhance healthcare assist in smart hospitals, and surgical phase recognition in laparoscopic videos is performed in ~\cite{cnnlstm,sar}. However, recognizing surgical activities in the OR has not been studied due to lack of data. 

Video action recognition has been studied extensively in computer vision domain due to its applications in areas such as surveillance, human-robot interaction, and human skill evaluation. While image representation architectures have matured relatively quickly in recent years, the same could not be said for those that deal with videos. This is mainly because videos occupy a significantly more complex space due to temporal dynamics. However, compilation of large-scale datasets such as Kinetics~\cite{kinetics400,kinetics600,kinetics700,sports1m} in the past few years has enabled researchers to explore deep architectures that take advantage of 3D convolution kernels leading to very deep, naturally spatio-temporal classifiers~\cite{i3d,c3d}. 

Action recognition methods fall under two categories depending on their input video. In action recognition on \textit{trimmed} videos, the goal is to classify a video that is trimmed such that it contains a single action~\cite{gsn,smn,vcc}. On the other hand, in an \textit{untrimmed} video there are two key differences: 1) one or several activities may be observed, 2) some parts of the video may contain no activity. Therefore, in the latter, the objective is to find the boundaries of each action as well as classify them in their correct category~\cite{mgg,tsr,emc}.

\vspace{-13pt}
\section{Dataset}
\vspace{-10pt}

The dataset used in this work has been captured from a single medical facility with two robotic ORs and da Vinci Xi systems. Institutional Review Board approvals have been obtained to enable our data collection. 
Two imaging carts were placed in each room, for a total of four carts, each equipped with two ToF cameras. The baseline between the cameras on each cart is 70 centimeters and their orientation is fixed. This results in slightly different view in videos captured by the cameras from the same cart. However, different carts in the same room are set in strategic positions, such that if a cart’s view is blocked due to clutter in the scene, the other cart can successfully capture the activities. With this setup, we captured 400 videos from 103 robotic surgical cases. Our dataset includes a variety of procedures performed by 8 surgeons/teams with different skill levels. A detailed breakdown of surgery types and surgeons is shown in  Figure~\ref{fig:surg}.


Following data collection, raw data from the ToF cameras was processed to extract 2D intensity frames which are then used to create videos. These videos were manually annotated into separate cases and their constituent surgical activities using a video annotation tool~\cite{via}. We do not perform annotation quality analysis and rely on expertise of one person closely familiar with the data. Although there are multiple ways to analyze the surgical process in a robotic OR, we focus on 10 activities (Fig.~\ref{fig:bar}) that are most relevant to robotic system utilization and non-operative workflows that are less investigated in prior studies. 


\begin{figure*}[t]
	\centering
	\includegraphics[width=0.9\linewidth]{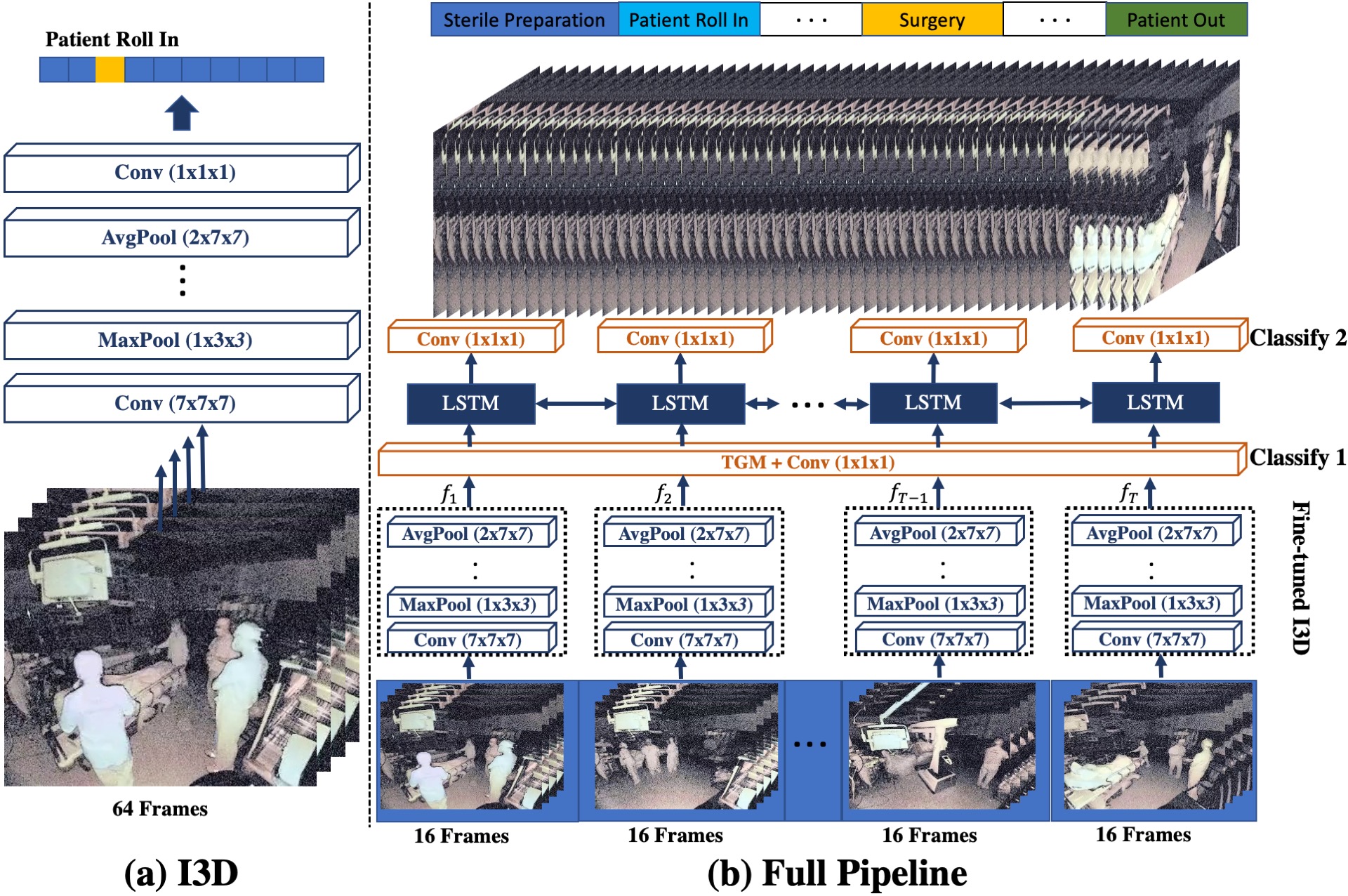}
	\vspace{-10pt}
	\caption{\scriptsize \textbf{(a)} Inflated 3D ConvNet (I3D). We fine-tune this network such that it can classify short clips (64/16-frames during training/testing respectively) sampled from our dataset. \textbf{(b)} In full pipeline, we break each video uniformly into 16-frame long clips and feed each to the fine-tuned I3D to extract features ($f_1, \cdots, f_T$). These features are then passed through TGM and convolution layers, a bidirectional LSTM, and a secondary 1D convolution layer that eventually predicts the action in each clips. Due to extreme length of the videos, using two classification layers facilitates convergence.}
	\label{fig:app} 
	\vspace{-20pt}
\end{figure*}
\vspace{-14pt}
\section{Methodology}
\vspace{-10pt}
In this section, we describe our model to perform activity recognition by breaking the problem into two subtasks; 1) extracting discriminative features for activity classes, and 2) using the features to recognize activities in full-length videos.

\vspace{-14pt}
\subsection{Inflated 3D ConvNet}
\label{sec:ar}
\vspace{-5pt}
Our first step to tackling full-length surgery video activity recognition is to extract discriminative features from the data. The intuition behind this task is to learn optimal representation for each class of activity such that the activities are easily distinguishable from one another. While various approaches exist, state-of-the-art models are built upon deep convolutional networks. Among them, the Inflated 3D ConvNet~\cite{i3d} (I3D) has become a reliable backbone to many video understanding methods, shown on the left panel of Figure~\ref{fig:app}. I3D is a very deep network consisting of several $3D$ convolution layers and Inception V1 modules, allowing it to extract discriminative spatio-temporal features from short clips. 



More formally, denote the videos as $\mathcal{V} = \{v_1,\cdots,v_N\}$ and set of activities as $\mathcal{C} = \{c_1,\cdots,c_K\}$ where $N$ and $K$ are the number of videos in the training and cardinality of set of activities respectively. Using the temporal annotations, each video can be partitioned into its corresponding activity segments, $v_i=\{s_{c_1},\cdots,s_{c_K}\}$ where $s_{c_i}$ represents the start and end times for activity type $c_i$. 

To fine-tune I3D, a short clip consisting of 64 consecutive frames is sampled from each timestamp $s_{c_i}$. This is done for every activity in every video in the training set, resulting in roughly $N*K$ total training samples. For activities that are longer than 64 frames, at every epoch, a different sample is selected (e.g., if a certain activity has $n$ frames, a random number between 0 and $n-63$ is chosen to serve as the starting point of the sample). This data augmentation technique enables better training. Once the training samples are decided, the model is trained via a categorical cross-entropy loss function $\mathcal{L} = \sum_{i=1}^{K} c_i \log p_i$, 
where $p_i$ is the probability of the model predicting class $c_i$ for the given sample. Once I3D is fine-tuned on our data, we use it as a backbone to extract spatio-temporal features from full-length videos. Thus, we first detach the classification layer from the network. Next, every video is uniformly segmented into 16-frame long clips and a 1024-d feature representation is extracted from each clip.

This network, once trained as described above, can be used to perform surgical activity recognition on full videos. To do so, we simply feed each video one clip (16 frames) at a time to obtain a classification decision per clip. We use this approach as a baseline and report its performance. As we expect (and later confirm through experiments), this method fails to perform reliably simply because classification of each clip is done locally; the model has no memory of previously seen clips that belong to the same video. Afterall, context must help in classifying a new clip. Another major drawback of this model is its inability to account for the order of the activities. We tackle these issues in the next section.

\vspace{-10pt}
\subsection{Full-Length Surgical Activity Recognition}
\label{lstm}
\vspace{-5pt}
In order to perform in-context prediction and model the inherent order of activities, we use a recent technique~\cite{tgm} and add a stack of three Temporal Gaussian Mixture \textbf{(TGM)} layers before an LSTM module to model local and global temporal information in the video. Each TGM layer learns several Gaussian (temporal) filters for each class of activity and uses them to provide local context before making a classification decision. The features extracted through this layer are concatenated with I3D features and are then fed to LSTM. LSTM processes the sequence in order and stores hidden states to serve as memory of its observation. These hidden representations are used to process the input at every time step and calculate the output, therefore making LSTM a powerful tool for the purpose of full-length surgical activity recognition.

After extracting features from the training videos, each video is then represented as $v_i=\{f_1,\cdots,f_T\}$ where $f_t$ is a feature extracted from $t^{\text{th}}$ clip and $T$ is the number of clips in $v_i$. At this stage of training, each video serves as a single training sample; the entire feature set of a video is fed to TGM layers before being processed by LSTM in sequential order to obtain $\mathcal{O}_i = \{o_1,\cdots,o_T\}$ where $o_t$ is the output after observing first $t$ features of the video, $o_t = LSTM(\{f_1,\cdots,f_t\})$.
This equation illustrates how LSTM represents each clip in a long video as a function of previous clips, hence providing context in representing each clip. This context helps to make classification decisions more robust. Furthermore, LSTM takes the order of activities into account as it processes the input sequentially. The output of the LSTM at every time step is fed to a convolution layer that serves as the classifier, classifying all the clips in the video.

To compare our model with closest work in the literature, Yengera et al.~\cite{cnnlstm} uses a \textbf{frame-level} CNN combined with an LSTM to perform surgical phase recognition on laparoscopic videos. Since the loss has to be backpropagated through time, using frame-level features (on such long videos) leads to extremely high space complexity, thus, resulting in convergence issues. Unlike them, we use clip level features, significantly reducing the complexity. Moreover, we further speed up convergence while preserving the generalization capability of the model by applying the classification loss before and after LSTM. This is achieved by feeding the TGM features to a 1D convolution layer with output dimension equal to number of activity classes (hence this layer can serve as a classifier on its own), and use its prediction as input to our LSTM module combined with a secondary classification layer. The same categorical cross-entropy is applied to the outputs of both classification layers and the network is trained end-to-end.
\begin{table*}[t]
	\centering
	\scriptsize
	\caption{\scriptsize{I3D fine-tuning results.}}
	\label{tab:i3d}
	\vspace{-10pt}
	\begin{tabular}{@{}lccccccccccc@{}}\toprule
		& \multicolumn{3}{c}{Random} & \phantom{abc}& \multicolumn{3}{c}{Procedure} &
		\phantom{abc} & \multicolumn{3}{c}{Surgeon}\\ \cmidrule{2-4} \cmidrule{6-8} \cmidrule{10-12}
		& Prec. & Rec. & F1 && Prec. & Rec. & F1 && Prec. & Rec. & F1\\ \midrule
		
		Sterile preparation & 97.0$\pm$1.4 & 96.3$\pm$1.7 & 96.5$\pm$1.3 && 86.44 & 78.46 & 82.26 && 81.08 & 88.24 & 84.51 \\
		Patient roll in     & 91.3$\pm$3.4 & 91.0$\pm$5.4 & 90.8$\pm$3.4 && 77.94 & 80.3 & 79.1 && 85.0 & 75.0 & 79.69 \\
		Patient preparation & 75.8$\pm$3.0 & 78.3$\pm$3.2 & 770.$\pm$2.6 && 55.56 & 78.57 & 65.09 && 61.97 & 62.86 & 62.41 \\ 
		Robot roll up       & 92.8$\pm$2.4 & 92.8$\pm$2.1 & 92.8$\pm$1.3 && 88.14 & 83.87 & 85.95 && 85.94 & 82.09 & 83.97 \\
		Robot docking       & 75.5$\pm$6.0 & 80.3$\pm$2.6 & 77.5$\pm$2.6 && 76.19 & 50.0 & 60.38 && 60.29 & 60.29 & 60.29 \\ 
		Surgery             & 80.5$\pm$1.7 & 86.0$\pm$5.4 & 83.0$\pm$1.4 && 69.51 & 81.43 & 75.0 && 73.33 & 74.32 & 73.83 \\
		Robot undocking     & 84.0$\pm$2.4 & 72.5$\pm$6.2 & 77.8$\pm$4.2 && 75.0 & 72.86 & 73.91 && 68.18 & 62.50 & 65.22 \\ 
		Robot roll back     & 96.3$\pm$2.2 & 90.8$\pm$3.3 & 93.3$\pm$2.4 && 94.64 & 80.30 & 86.89 && 98.08 & 76.12 & 85.71 \\
		Patient close       & 77.8$\pm$1.0 & 79.3$\pm$2.2 & 78.5$\pm$0.6 && 69.44 & 69.44 & 69.44 && 67.16 & 62.5 & 64.75 \\
		Patient roll out    & 88.5$\pm$3.4 & 90.3$\pm$4.9 & 89.3$\pm$1.5 && 73.53 & 73.53 & 73.53 && 63.64 & 90.0 & 74.56 \\
		\bottomrule
	\end{tabular}
	\vspace{-20pt}
\end{table*}
\vspace{-15pt}
\section{Experimental Setup and Result Analysis}
\vspace{-10pt}
\paragraph{\textbf{Train/test Data Split.}} In order to comprehensively study the generalization power of the model, we perform experiments using different data split schemes.

\noindent \textbf{I.} \textit{Random split.} In this experiment, we randomly select 80\% of the videos for training and the rest for testing. We do not take precautions to guarantee that all 4 videos belonging to the same case are either in train or test sets. This experiment is designed to find a meaningful upper bound to model’s performance. We repeat this several times and report the mean and standard deviation of the performance to ensure an unbiased evaluation.

\noindent \textbf{II.} \textit{Operating room split.} Since the entirety of our data is captured in two ORs within the same medical center, we can assess the model's generalization capability in cross-room train-test split. We train on videos from OR1 (57.5\% of all the videos) and test the model on videos captured in OR2. We expect to see a drop in performance as the model is trained with less data.

\noindent \textbf{III.} \textit{Procedure type split.} With this split, we assess if the model can generalize to novel procedures. In other words, we evaluate the model's performance by training on certain procedure types (Cholecystectomy, etc.) and test it on unseen categories. We used 12 out of 28 types of surgeries covered in our dataset for testing and the remaining for training. This results in a 80\%-20\% train-test split.

\noindent \textbf{IV.} \textit{Surgeon split.} Surgeons lead the procedure and significantly affect the surgical workflow. In this experiment, we train our model on videos from select surgeons and assess the model's performance on videos from the remaining surgeons. The surgeons are split such that we obtain a 80\%-20\% train-test split.


\vspace{-10pt}
\paragraph{\textbf{Evaluation Metrics.}}
I3D network is designed to classify short video clips and is evaluated using precision, recall, and F1 score. The full pipeline is evaluated similar to untrimmed action recognition algorithms using mean Average Precision (mAP).



\begin{table*}[t]
	\centering
	\scriptsize
	\caption{\scriptsize Full-length surgical activity recognition results (mAP).}
	\vspace{-5pt}
	\label{tab:map}
	\begin{tabular}{@{}lccccc@{}}\toprule
		& \multicolumn{1}{c}{I3D} & &
		\multicolumn{1}{c}{I3D-LSTM} & &
		\multicolumn{1}{c}{I3D-TGM-LSTM}\\
        \midrule
        Random          & 62.16 $\pm$ 1.60 && \textbf{88.81 $\pm$ 1.30} && 88.75 $\pm$ 2.07 \\
        Procedure    & 47.98 && 75.39 && \textbf{78.04} \\
        Surgeon      & 50.31 && 78.29 && \textbf{78.77} \\
		\bottomrule
	\end{tabular}
	\vspace{-15pt}
\end{table*}
\vspace{-10pt}
\paragraph{\textbf{Fine-Tuned I3D Results.}}
Per class precision, recall, and F1-scores of our experiments are shown in Table~\ref{tab:i3d}. Following a random data split scheme, training I3D results in a robust clip-level action recognition network. We performed 4 rounds of experiments, every time with a new random data split. The model is able to maintain high average scores while keeping the standard deviation low.

When separating videos based on which OR they were captured in, the performance drops significantly (precision: 64.9, recall: 61.9, F1: 61.8\% on average). The reason behind this observation is twofold: 1) the number of training videos is significantly less, 2) certain types of operations (e.g., lung Lobectomy, Thymectomy) had no instance in OR1 videos and were only seen in OR2. Thus, our clip recognition model is less robust compared to random split experiments.

Next, we test the model's generalization power when dealing with novel procedure types. This split is more difficult compared to random as the testing surgery types were never seen during the training. As shown in Table~\ref{tab:i3d}, F1-scores drop in every class of activity, yet retain an average F1 score of 75.15\%. 

Finally, when splitting the videos based on the surgeons, we observe the same trend as procedure type split. This confirms that I3D is able to maintain a decent performance when tested on unseen surgeons.
\vspace{-8pt}
\paragraph{\textbf{Full-Length Surgical Activity Recognition Results.}}
As described in section \ref{sec:ar}, we can use the baseline I3D to perform activity recognition on full-length videos. To do so, we add a 1d convolution on top of the pretrained I3D to train (only the convolution layer) on full-length videos. The performance of this baseline, in mean Average Precision (mAP), is reported under the first column of Table~\ref{tab:map}. Not surprisingly, the trend is similar to our clip-level recognition experiments; the performance is higher when using more training videos. The base I3D yields mAP of 36.91\% when we split the videos by the OR they were captured in. We do not report this number in the table simply because the train-test split ratio is significantly different than the other 3 data split schemes.

When comparing I3D to our full pipeline (I3D-LSTM), we see a significant boost in performance under every split scheme. This is expected since the classification of clips is done within the context of other clips in the video, and the order of activities is learned. Peak performance of \textbf{88.81\%} mAP on average was obtained. In worst case scenario (OR data split), I3D-LSTM reaches 58.40 mAP, which is significantly higher than the base I3D.

Finally, we see that TGM layer is increasing the model's generalization capability; it performs on-par with I3D-LSTM on random split, but outperforms it on all 3 other splits. I3D-TGM-LSTM reaches mAP of $63.47\%$ on OR split, outperforming I3D-LSTM by a margin of 5\%. As mentioned in previous sections, it is important to classify each clip in the video with consideration of its neighboring clips. This is the underlying reason why I3D-TGM-LSTM reaches state-of-the-art performance. The temporal filters pool class-specific features from neighboring clips. These features provide complimentary information to LSTM unit, resulting in overall increased performance.

\vspace{-10pt}
\section{Conclusion and Future Work}
\vspace{-10pt}
In this paper, we introduce the first large-scale robot-assisted operation dataset with 400 full-length videos captured from real surgery cases. We temporally annotate every video with 10 clinically relevant activities. Furthermore, we use this dataset to design a model to recognize activities throughout the video. To this end, we fine-tune the Inflated 3D ConvNet model on our dataset to learn discriminative feature representations for different types of activities. Next, using Temporal Gaussian Mixture layers and a Long Short Term Memory unit, we enable the model to keep track of activities, thus learning their order.

There are several directions for expanding our efforts in this paper. From the data perspective, we are in the process of acquiring videos from different medical centers. This allows us to: 1) assess the model's generalization ability more comprehensively, 2) explore deeper architectures. From the application stand point, our proposed framework can be used in future to provide automated workflow and efficiency metrics to surgical teams and medical institutions.

\end{document}